\documentclass[letterpaper, 10 pt, conference]{ieeeconf}  

\IEEEoverridecommandlockouts                              
\usepackage{cite}
\usepackage{amsmath,amssymb,amsfonts}
\usepackage{algorithmic}
\usepackage{graphicx}
\usepackage{textcomp}
\usepackage{multirow}
\usepackage{xcolor}
\title{\LARGE \bf
Zero-Shot Object Goal Visual Navigation With Class-Independent Relationship Network
}

\author{Albert Author$^{1}$ and Bernard D. Researcher$^{2}$
\thanks{*This work was not supported by any organization}
\thanks{$^{1}$Albert Author is with Faculty of Electrical Engineering, Mathematics and Computer Science,
        University of Twente, 7500 AE Enschede, The Netherlands
        {\tt\small albert.author@papercept.net}}%
\thanks{$^{2}$Bernard D. Researcheris with the Department of Electrical Engineering, Wright State University,
        Dayton, OH 45435, USA
        {\tt\small b.d.researcher@ieee.org}}%
}

\author{Xinting Li$^{1}$, Shiguang Zhang$^{2*}$, Yue LU$^{1}$, Kerry Dang$^{1}$, Lingyan Ran$^{1}$ 
\thanks{$^{1}$Xinting Li, Yue LU, Kerry Dang, and Lingyan Ran is with Faculty of Computer Technology, School of Computer Science, Northwestern Polytechnical University. $^{2}$Shiguang Zhang is with the Chinese People's Liberation Army Unit 32620. (* denotes the corresponding author:
        {\tt\small 13709759484@139.com} )}%
}

\begin{document}

\maketitle
\thispagestyle{empty}
\pagestyle{empty}

\begin{abstract}

This paper investigates the zero-shot object goal visual navigation problem. In the object goal visual navigation task, the agent needs to locate navigation targets from its egocentric visual input. "Zero-shot" means that the target the agent needs to find is not trained during the training phase. To address the issue of coupling navigation ability with target features during training, we propose the Class-Independent Relationship Network (CIRN). This method combines target detection information with the relative semantic similarity between the target and the navigation target, and constructs a brand new state representation based on similarity ranking, this state representation does not include target feature or environment feature, effectively decoupling the agent's navigation ability from target features. And a Graph Convolutional Network (GCN) is employed to learn the relationships between different objects based on their similarities. During testing, our approach demonstrates strong generalization capabilities, including zero-shot navigation tasks with different targets and environments. Through extensive experiments in the AI2-THOR virtual environment, our method outperforms the current state-of-the-art approaches in the zero-shot object goal visual navigation task. Furthermore, we conducted experiments in more challenging cross-target and cross-scene settings, which further validate the robustness and generalization ability of our method. Our code is available at: https://github.com/SmartAndCleverRobot/ICRA-CIRN.

\end{abstract}


\section{INTRODUCTION}
Visual navigation is a fundamental problem in robotics and artificial intelligence. Object goal visual navigation tasks aim to command agents to navigate to specific targets in 3D environments using egocentric camera views. In recent years, visual navigation has garnered significant attention in the fields of artificial intelligence and computer vision, showing extensive applications in areas such as automated home services, warehouse management, and the hospitality industry.

Due to the success of reinforcement learning based methods in robotic tasks, numerous reinforcement learning-based approaches for map-less visual navigation have emerged\cite{ref7,ref8,ref9,ref10,ref11}. Unlike traditional map-based visual navigation methods, these recent reinforcement learning based approaches are end-to-end solutions, directly mapping visual information and navigation goals to actions. Consequently, they require minimal manual engineering and serve as the foundation for the new generation of AI-driven visual navigation tasks.

Currently, though researchers have achieved promising results in target navigation, these methods possess significant limitations as they are tested on the same targets they were trained on. This means that an agent needs to be trained on how to find a specific class before it can locate that class during testing. However, in the real world, there are countless classes, making it infeasible to train the agent on each one. Thus, the investigation of Zero-Shot Object Goal Visual Navigation (ZSON) tasks becomes necessary, where agents are expected to find untrained classes in the testing environment.

In previous works, it was common to combine the features of navigation targets with the current visual information as the state for reinforcement learning. While this design allows agents to have a clear understanding of the navigation targets, it also leads to agents learning both the features of navigation targets and how to navigate to those targets. Specifically, this approach couples the agent's navigation ability with the features of navigation targets during training. Consequently, during testing, when the agent is tasked to find a new target, the combination of the new target and the current visual information forms a completely new state, making it hard for the agent to comprehend its navigation target. Therefore, we believe that the key to solving the ZSON task is to decouple the agent's navigation ability and the feature of the navigation target during training, and try to make the agent learn pure navigation ability rather than the ability to navigate to specific targets which trained in the training stage.

\begin{figure*}
\includegraphics[width=0.75\linewidth]{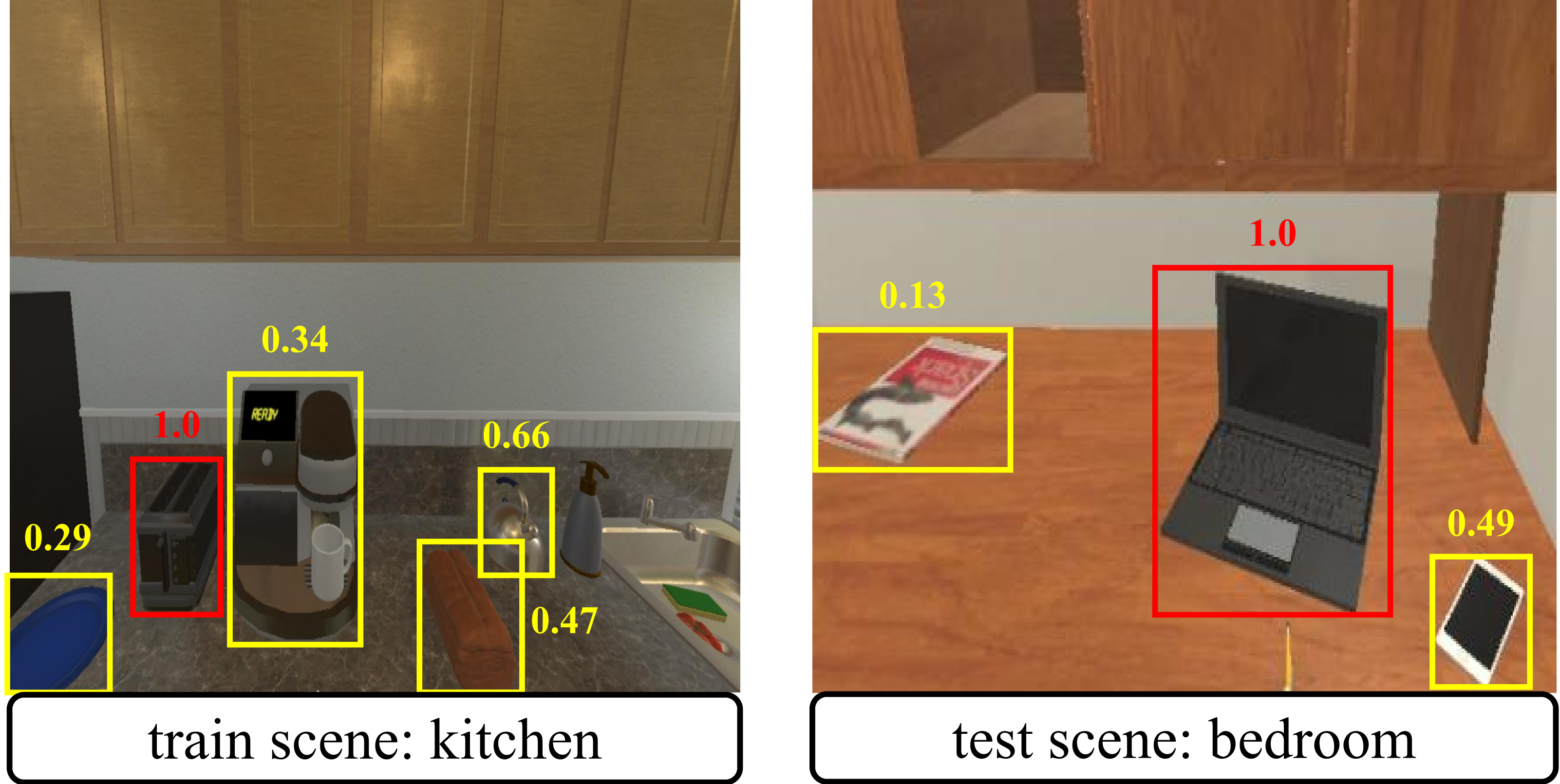}
\centering
\caption{The motivation of this method is to decouple the navigation ability of the agent from the navigation target, as shown in the above figure. The red marked object in the figure represents the navigation target, the yellow marked other object, the numbers on the bounding box refer to the semantic similarity between this object and the navigation target, the navigation target in the left figure is the toaster, and the navigation target in the right figure is the Laptop. Although the test environment and the test target are different from the training, However, in the state set by CIRN, is not contain the feature of the object or environment, and the difference between the two is only in spatial location and semantic similarity.} \label{liucheng}
\end{figure*}

Inspiration from\cite{ref15} we believe that for navigation tasks, target detection information is sufficient for the agent to locate the spatial position of the target. On the issue of how to represent the object class, we prefer using a relative class compared to using the object's word embedding\cite{ref8} or the visual feature\cite{ref9} to represent their objects. Because both word embedding and visual feature contain distinctive feature that uniquely identify the navigation target and our goal is to eliminate these feature belonging to a specific class from the state representation. Therefore, we use the relative semantic similarity between objects and navigation targets to represent different objects. We then sort the information of multiple objects in descending order based on their relative semantic similarity. Compared to the preset number of object categories and the fixed position of categories in the state matrix\cite{ref15}, our method can be applied to all scenarios, regardless of the specific class. Consequently, the sole identifier distinguishing objects is their semantic similarity to the navigation target, regardless of their class. Therefore, we propose the Class-Independent Relationship Network (CIRN). As shown in Fig. 1, the key idea of CIRN is to ensure that the information received by the agent does not undergo significant changes due to variations in navigation targets and scenes. Specifically, CIRN uses object detection information and semantic similarity with navigation targets as state, effectively decoupling the feature information of the navigation target from the navigation decision-making process. This enables the agent to navigate more flexibly and accurately when facing new navigation targets.

We conducted an extensive evaluation of our method in the AI2-THOR virtual environment, which includes 120 indoor scenes and 22 commonly targeted classes. Our results demonstrate a remarkable improvement compared to the current state-of-the-art methods. Moreover, even under more challenging experimental settings, we consistently maintain a high success rate, providing compelling evidence that CIRN is a highly effective approach for tackling the ZSON task.

\begin{figure*}
\includegraphics[width=0.75\linewidth]{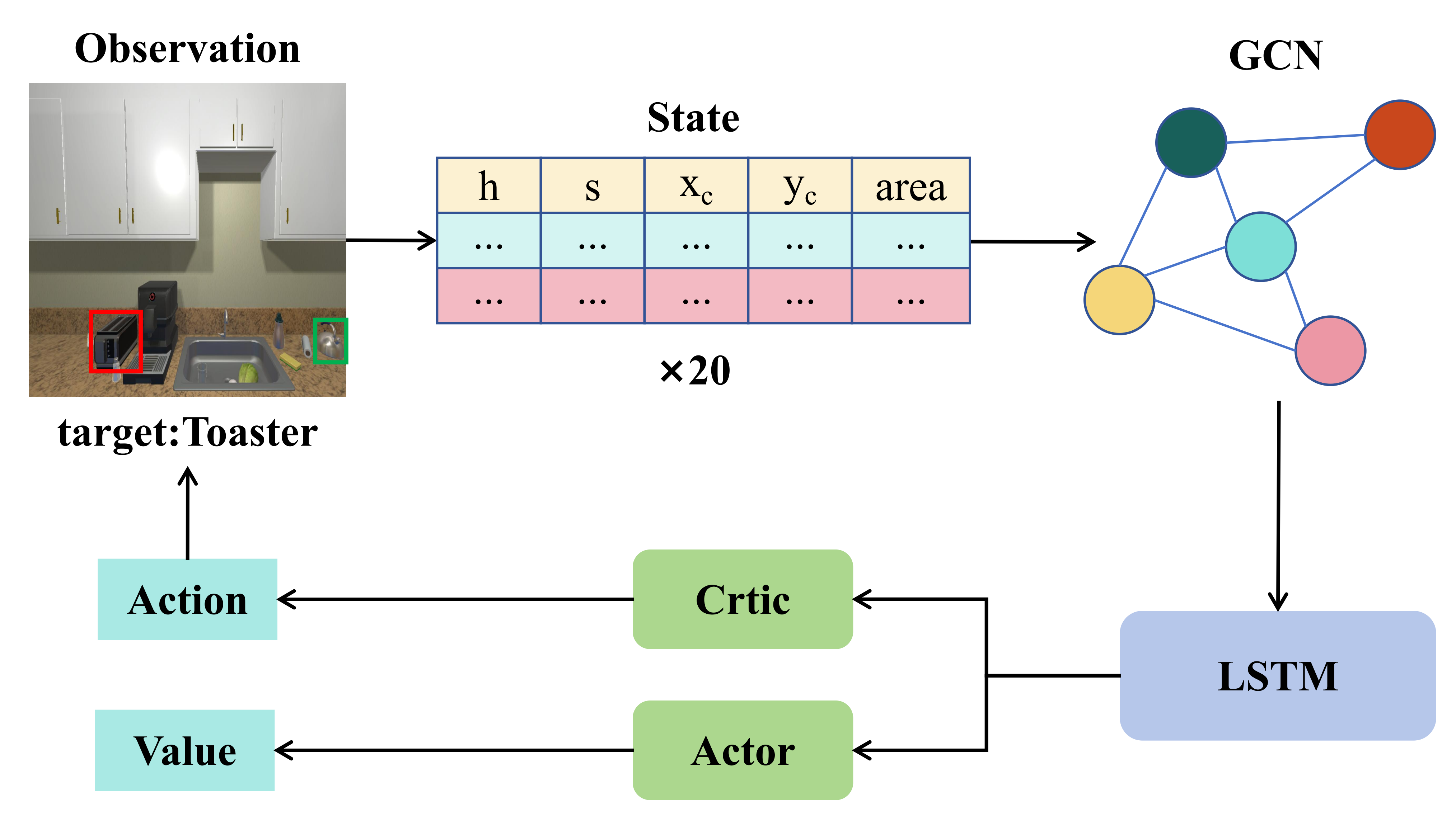}
\centering
\caption{Architecture Overview: The model receives input consisting of object detection information for different classes within the field of view, as well as their semantic similarity to the target, sorted in descending order according to semantic similarity. This input processes through the GCN module before being fed into the LSTM network. The LSTM module is responsible for extracting and retaining past action information. Ultimately, an actor-critic network is used to generate actions.} \label{liucheng}
\end{figure*}

\section{ RELATED WORK}

Visual navigation is a fundamental challenge in the field of mobile robotics. Traditional navigation methods often rely on environmental maps and involve a multi-step process, including mapping, localization, and path planning \cite{ref1,ref2,ref3,ref4,ref5,ref6}.  With the development of reinforcement learning, and reinforcement learning-based navigation methods have gained popularity due to their ability to tackle complex tasks through end-to-end approaches. Since the pioneering work of Zhu et al. \cite{ref7}, who introduced an end-to-end navigation model based on deep learning, incorporating localization, mapping, exploration, and semantic recognition, target-driven visual navigation has seen rapid advancements, leading to the proposal of several efficient models.Wortsman et al. \cite{ref8} proposed a meta-learning based approach that dynamically adjusts the navigation policy to adapt to changes in the environment, and the agent  learns self-supervised interaction losses to enhance navigation effectiveness. Lee et al. \cite{ref9} introduced object relation graphs to learn spatial relationships between classes appearing in navigation, thereby providing better guidance to agents during navigation. To further improve information representation, they proposed the novel Visual Transformer network (VTNet) \cite{ref10}, which extracts informative feature representations in navigation. These representations not only encode object relationships but also establish a strong correlation with the navigation signal, thereby aiding the agent in making informed decisions for its next actions. Bar et al \cite{ref12}. proposed an attention mechanism that combines navigation information and navigation actions to better help the agent find the target.

However, existing visual navigation methods have significant limitations because models trained using the above-mentioned approaches can only recognize navigation target classes that have been previously seen during training. It is impractical to train models for every possible class in the real world, as there are countless classes. Therefore, Zero-Shot Object Navigation (ZSON) is of great practical significance. ZSON refers to the agent's ability to extend its navigation capabilities to new, unseen classes after learning from a set of navigation target classes. Apoorv Khandelwal et al. [13] first introduced the concept of ZSON and designed a method based on the CLIP [14] model, which utilizes CLIP's zero-shot capability to determine whether the navigation target is in the agent's field of view, guiding its decision-making process. Samir Yitzhak Gadre et al. [14], based on [13], used the Grad-CAM heatmaps generated by CLIP as the state representation of the target's location in the image and developed a deep map-based navigation system.Qianfan Zhao [15] proposed a method based on semantic similarity, which uses the semantic similarity between the test navigation target and the trained target to guide the agent's navigation. Arjun Majumdar [16] introduced a method called Multimodal Goal Embeddings, which leverages the semantic similarity between images and text in the CLIP model to train on the ImageNav task and test on the objectNav task. Existing methods mainly focus on exploiting the similarity between test and train targets, as seen in the method proposed in [15], where the test results depend on the number of target classes during training. However, we believe that using the navigation target's features as the state during training is the key reason why previous methods perform poorly when faced with new targets. This coupling of learned navigation capabilities with the navigation targets during training greatly hinders the model's generalization ability.

Therefore, we propose a method called Class-Independent Relationship Network (CIRN), which successfully decouples the navigation capability from the navigation targets and achieves optimal performance across various tasks.

\section{METHOD}

\subsection{Task Definition}\label{AA}
A navigation task comprises a scene $S$, an initial point $p$, and a navigation target $t$. The primary objective of the agent is to locate the navigation target $t$ within the 3D environment from its starting position, using a limited number of steps. The agent's action space consists of six actions: {MoveAhead, RotateLeft/RotateRight, LookDown/LookUp, Done}. We learn a policy $\pi_\theta(a|s)$ At each step, the agent receives an egocentric RGB image $s$ from the scene and information about the target object $o$, based on which it selects an action $a$ from the action space. We use policy gradient to improve the model's parameters $\theta$. The update direction of this parameter is to maximize the expected reward on a sequence of actions in a given episode. We also use the actor-critic algorithm to minimize the navigation loss $\mathbb{L}_{nav}(\theta , a)$, which uses the critic to generate the policy and the actor to generate the value function. The navigation task is considered successfully completed if the agent executes the "done" action when it is within 1.5 meter of an instance of the target object class and the object is within the agent's field of view. A collection of all the steps from the beginning to the end of a task is called an episode. Regarding the setting of the reward function, we follow the approach in \cite{ref15}. After the agent successfully completes a task, it is rewarded with a positive reward of 5. Furthermore, the reward increases as the cosine similarity of the objects within the agent's field of view becomes higher. On the other hand, if there are no objects present in the field of view, the agent receives a reward of -0.01.

The scene $S$ is divided into test scenes and training scenes, and the navigation targets are categorized into test targets and training targets. During the training phase, the agent uses training targets as navigation objectives within the training scenes. On the other hand, during the testing phase, the agent utilizes test targets as navigation objectives within the test scenes.

\begin{table*}[htbp]
\caption{The result of the other methods and ours in testing scenes.}
\begin{center}
\begin{tabular}{|c|c|c|c|c|c|c|c|c|c|}
\hline 
\multirow{3}*{model} & \multirow{3}*{train/test split} & \multicolumn{4}{c|}{test class}  & \multicolumn{4}{c|}{train class}  \\  
\cline{3-10}
& & \multicolumn{2}{c}{$L\ge1$} & \multicolumn{2}{|c}{$L\ge5$} & \multicolumn{2}{|c}{$L\ge1$} & \multicolumn{2}{|c|}{$L\ge5$} \\
\cline{3-10}
& & SR & SPL & SR & SPL & SR & SPL & SR & SPL \\ \hline
random & 18/4 & 10.8 & 2.1& 0.9& 0.3&9.5&3.3& 1.0& 0.4 \\ \hline
SSNet\cite{ref15} & 18/4 & 28.6 & 9.0 & 12.5 & 5.6 & 59.0 & 19.7 & 38.6 & 18.3 \\ \hline
\textbf{CIRN} & 18/4 & \textbf{66.9} & 28.5 & 47.7 & 24.9 & 63.4 & 26.5 & 41.7 & 24.0 \\ \hline
random & 14/8 & 8.2 & 3.5& 0.5& 0.1&8.9&3.0& 0.5& 0.3 \\ \hline
SSNet\cite{ref15} & 14/8 & 21.5 & 7.0 & 13.0 & 6.7 & 59.3 & 24.5 & 35.2 & 19.3 \\ \hline
\textbf{CIRN} & 14/8 & \textbf{64.4} & 25.7 & 41.1 & 21.9 & 64.8 & 26.1 & 44.6 & 24.2 \\ \hline
\end{tabular}
\label{tab1}
\end{center}
\end{table*}
\subsection{Model Architecture}
There is a large amount of navigation-irrelevant information in visual data, such as color, texture, and so on, which can cause the model to overfit to the environment or specific objects. Inspired by \cite{ref15}, it has been shown that using only object detection information can still complete the visual navigation task very well. Therefore, to eliminate differences in visual information across different environments and objects, we exclude any visual data from our state representation.The architecture of the proposed CIRN is shown in Fig. 2. Our state is defined as a 20x5 matrix, where each row can be represented as $r_i = [h, s, x_c, y_c, area]$ of object $i$. The first element $h$ indicates whether the object is the target; $h=1$ if it is, otherwise $h=0$. The second element $s$ represents the semantic similarity between the object and the navigation target. The third and fourth elements, $x_c$ and $y_c$, denote the center coordinates of the detection box, and the fifth element, area, represents the area of the detection box. We sort the rows based on the value of $s$ in descending order and select the top twenty object detection boxes. If the number of object detection boxes under the current visual data is less than twenty, the remaining rows are set to zero. Our work mainly focuses on policy learning, so we directly use the ground truth detection results, which is the same as other works \cite{ref8,ref17,ref18}.

The semantic similarity in the state is obtained by calculating the cosine similarity (CS) of the word embedding between the object and the target, and the calculation formula is as follows:
\begin{equation}
CosineSimilarity(g_i,g_t)=\frac{g_i \cdot g_t}{|g_i|*|g_t|}\label{eq}
\end{equation}

Where $g_i$ represents the word embedding of object $i$, and $g_t$ represents the word embedding of navigation target.It can be observed that the state information we propose does not contain any visual information. The spatial location information of objects in the state is used to identify their positions, while semantic similarity is employed to differentiate between different objects. Additionally, the object's class information is relative to the navigation target; if an object has a high semantic similarity to the target, its order is prioritized. As a result, our information does not include any specific object details. Thus, during testing, when faced with a new navigation target, the information contained in the state remains indistinguishable from that during training.

After constructing the state, we utilize a Graph Convolutional Network (GCN) to learn the relationships between objects based on their CS arrangements. In navigation task, the state information can be regarded as a graph, where each object corresponds to a node, and the semantic similarity and spatial position information constitute the node features. The GCN performs multiple layers of convolutional operations, gradually updating and integrating the features of each node, enabling the node representations to encapsulate both global and local semantic information. The role of the Graph Convolutional Network lies in considering the relationships and interactions between objects within the state, thereby enhancing the expressive power of the state representation. By learning the connectivity patterns between nodes in the graph, the model can capture the semantic connections and spatial relationships between objects, which is particularly crucial for navigation tasks as they require considering the layout and features of the surrounding environment, not just individual object information.  Subsequently, we input the state processed by the GCN into the policy network (which includes LSTM and Actor-Critic network). The policy network then outputs the corresponding policy $\pi$ for the current state.

\begin{table}[htbp]
\caption{train in kitchen and test in bedroom}
\begin{center}
\begin{tabular}{|c|c|c|c|c|c|c|c|c|}
\hline 
\multirow{3}*{model} & \multicolumn{4}{c|}{test class}  & \multicolumn{4}{c|}{train class}  \\  
\cline{2-9}
& \multicolumn{2}{c}{$L\ge1$} & \multicolumn{2}{|c}{$L\ge5$} & \multicolumn{2}{|c}{$L\ge1$} & \multicolumn{2}{|c|}{$L\ge5$} \\
\cline{2-9}
 & SR & SPL & SR & SPL & SR & SPL & SR & SPL \\ \hline
SSNet\cite{ref15} & 3.6 & 0.68 & 0.0 & 0.0 & 78.4 & 32.8 & 61.5 & 32.9 \\ \hline
\textbf{CIRN} & \textbf{46.4} & 20.3 & 24.5 & 14.3 & 84.0 & 31.0 & 71.5& 34.9 \\ \hline
\end{tabular}
\label{tab1}
\end{center}
\end{table}

\begin{table}[htbp]
\caption{train in kitchen and test in living room}
\begin{center}
\begin{tabular}{|c|c|c|c|c|c|c|c|c|}
\hline 
\multirow{3}*{model} & \multicolumn{4}{c|}{test class}  & \multicolumn{4}{c|}{train class}  \\  
\cline{2-9}
& \multicolumn{2}{c}{$L\ge1$} & \multicolumn{2}{|c}{$L\ge5$} & \multicolumn{2}{|c}{$L\ge1$} & \multicolumn{2}{|c|}{$L\ge5$} \\
\cline{2-9}
 & SR & SPL & SR & SPL & SR & SPL & SR & SPL \\ \hline
SSNet\cite{ref15} & 10.8 & 2.2 & 2.5 & 1.0 & 78.4 & 32.8 & 61.5 & 32.9 \\ \hline
\textbf{CIRN} & \textbf{54.4} & 23.3 & 29.4 & 17.5 & 84.0 & 31.0 & 71.5& 34.9 \\ \hline
\end{tabular}
\label{tab1}
\end{center}
\end{table}

\section{EXPERIMENT}
We train and test using the AI2-THOR virtual environment, which comprises scenes from four categories: kitchen, living room, bedroom, and bathroom. Each type of scene has 30 rooms, 20 of which are divided into training scenes and 10 are divided into test scenes. 

We follow the evaluation metrics as described in references \cite{ref7,ref9,ref10}, which include the Success Rate (SR) and Success weighted by Path Length (SPL). SR is defined as $\frac{1}{N}\sum_{i=1}^{N}S_i$, while SPL is defined as $\frac{1}{N}\sum_{i=1}^{N}S_i\frac{L_i}{\max(P_i,L_i)}$. where N is the number of episodes, $S_i$ is a binary indicator of success in episode $i$, where $P_i$ is path length and $L_i$ is the length of the optimal path length of the agent in an episode. Since the behavior of the agent's policy is different for short and long paths. We also refer to trajectories with an optimal path length of at least 5, denoted by L $\ge$ 5 (L is the optimal episode length).

The experimental setup in Table 1 follows that of \cite{ref15}. We trained our model on four different scenes, totaling 80 train scenes, and tested it on the remaining 40 test environments. Among them, the 22 navigation targets were divided into 18/4 and 14/8. Table 1 presents a comparison between our experimental results and those of \cite{ref15}. The results show that our method performs exceptionally well, even surpassing the success rate on the test class compared to the train class. This indicates that our approach successfully decouples the navigation capability from the navigation target, achieving remarkable zero-shot ability on similar scene types.

In order to better demonstrate the zero-shot capability of our method and at the same time demonstrate the superiority of our method, we make more challenging improvements to the experimental setup. We trained in the kitchen and living room respectively, and then tested in Bedroom and bathroom. This means that the tested scene and target will not be seen by the agent at all during training, which is more challenging because it is cross-target, cross-scene, and different from the experimental setting above. This experimental setting has never seen a target in the training phase, so it better reflects the zero-shot ability of the network and also tests the model's broader generalization ability. Table 2 and Table 3 are the results of the agent training in the kitchen and testing in the bedroom and living room. It can be seen that compared with the train class in Table 3 and Table 4, the success rate has not yet reached the level of the train class, because it is in an unfamiliar environment after all. , but compared with the previous method, it is a big improvement, indicating that our model has a strong zero-shot ability, and can find unseen targets in different types of scenes.

Table 3 and Table 4 are the test results of the agent trained in the living room in the bathroom and kitchen. It can be seen that the test results in the bathroom have a very high success rate, mainly because the bathroom space is small and the layout is simple. At the same time, you can see the value of his SPL is not high, indicating that the agent will first explore the environment in an unfamiliar environment. The success rate of Agent in the kitchen is about $20\%$ lower than that of the train class, indicating that our method still has room for improvement in complex environments.

\begin{table}[htbp]
\caption{train in living room and test in bathroom}
\begin{center}
\begin{tabular}{|c|c|c|c|c|c|c|c|c|}
\hline 
\multirow{3}*{model} & \multicolumn{4}{c|}{test class}  & \multicolumn{4}{c|}{train class}  \\  
\cline{2-9}
& \multicolumn{2}{c}{$L\ge1$} & \multicolumn{2}{|c}{$L\ge5$} & \multicolumn{2}{|c}{$L\ge1$} & \multicolumn{2}{|c|}{$L\ge5$} \\
\cline{2-9}
 & SR & SPL & SR & SPL & SR & SPL & SR & SPL \\ \hline
SSNet\cite{ref15} & 8.8& 1.5 & 3.5 & 1.0 & 48.4 & 17.3 & 26.2 & 13.8 \\ \hline
\textbf{CIRN} & \textbf{80.4} & 15.6 & 76.7 & 19.8 & 58.8 & 22.2 & 44.1 & 21.3 \\ \hline
\end{tabular}
\label{tab1}
\end{center}
\end{table}

\begin{table}[htbp]
\caption{train in living room and test in kitchen}
\begin{center}
\begin{tabular}{|c|c|c|c|c|c|c|c|c|}
\hline 
\multirow{3}*{model} & \multicolumn{4}{c|}{test class}  & \multicolumn{4}{c|}{train class}  \\  
\cline{2-9}
& \multicolumn{2}{c}{$L\ge1$} & \multicolumn{2}{|c}{$L\ge5$} & \multicolumn{2}{|c}{$L\ge1$} & \multicolumn{2}{|c|}{$L\ge5$} \\
\cline{2-9}
 & SR & SPL & SR & SPL & SR & SPL & SR & SPL \\ \hline
SSNet\cite{ref15} & 8.8& 1.6 & 1.3 & 0.4 & 48.4 & 17.3 & 26.2 & 13.8 \\ \hline
\textbf{CIRN} & \textbf{66.8} & 14.0 & 50.3 & 15.3 & 58.8 & 22.2 & 44.1 & 21.3 \\ \hline
\end{tabular}
\label{tab1}
\end{center}
\end{table}

\section{CONCLUSION}
In this paper, we address the issue of coupling between navigation capability and navigation target features in Zero-Shot Object Goal Visual Navigation (ZSON) by proposing the Class-Independent Relationship Network (CIRN). The CIRN network utilizes object detection information and relative semantic similarity, sorted based on the semantic similarity, to construct a novel state representation. This state representation does not contain specific target or environmental features, effectively decoupling the agent's learning of navigation capability from navigation target features. As a result, our method maintains robust performance across various test conditions, such as cross-target and cross-scene.

Since we incorporate object detection information as part of our state representation, the zero-shot capability of our method depends on the detection range of the object detector. With a more advanced object detector, the zero-shot scope of our method can be extended.In future work, we aim to apply our model to physical robot platforms to evaluate its performance under real-world conditions.







\end{document}